%% file: main.tex
\documentclass[a4paper]{article}

\newcount\Comments 
\newcommand{\kibitz}[2]{\ifnum\Comments=1\textcolor{#1}{#2}\fi}
\usepackage{xcolor}

\newcommand{\partitle}[1]{\smallskip \noindent \textbf{#1.}}

\usepackage{graphicx}
\usepackage{makecell} %
\usepackage{mathtools}
\usepackage{amsthm}
\usepackage{indentfirst}
\usepackage{adjustbox}
\usepackage{caption}
\usepackage{subcaption}
\usepackage[ruled,vlined]{algorithm2e}
\usepackage[noend]{algpseudocode}
\usepackage{amssymb}
\newlength\myindent
\newcommand{\method}{CrowdTeacher}

\theoremstyle{definition}
\newtheorem{definition}{Definition}

\begin{document}
\title{\method: Robust Co-teaching with Noisy Answers \& Sample-specific Perturbations for Tabular Data}
\author{Mani Sotoodeh\footnote{{Emory University, Atlanta GA , USA }{Email: msotood@emory.edu}}\and
Li Xiong\and
Joyce Ho }

\maketitle              
\begin{abstract}
Samples with ground truth labels may not always be available in numerous domains. 
While learning from crowdsourcing labels has been explored, existing models can still fail in the presence of sparse, unreliable, or differing annotations.
Co-teaching methods have shown promising improvements for computer vision problems with noisy labels by employing two classifiers trained on each others' confident samples in each batch. Inspired by the idea of separating confident and uncertain samples during the training process, we extend it for the crowdsourcing problem.
Our model, \method, uses the idea that perturbation in the input space model can improve the robustness of the classifier for noisy labels. Treating crowdsourcing annotations as a source of noisy labeling, we perturb samples based on the certainty from the aggregated annotations. The perturbed samples are fed to a Co-teaching algorithm tuned to also accommodate smaller tabular data. We showcase the boost in predictive power attained using \method~for both synthetic and real datasets across various label density settings. Our experiments reveal that our proposed approach beats baselines modeling individual annotations and then combining them, methods simultaneously learning a classifier and inferring truth labels, and the Co-teaching algorithm with aggregated labels through common truth inference methods.

\end{abstract}

\input{intro.tex}

\input{method.tex}

\input{experiment.tex}
\input{result.tex}

\input{conclusion.tex}

\bibliographystyle{abbrv}
\bibliography{references.bib}

\end{document}

%% file: intro.tex
\section{Introduction and Background}
Labeled data is essential to guarantee the success of increasingly more complex classifiers.
Unfortunately obtaining large quantities of high-quality labels can be cost-prohibitive for several fields.
For example, in the medical domains, it may take a clinician several hours to annotate the health records of hundreds of patients.
One alternative is to gather labels using crowdsourcing, where remotely located workers are utilized to perform the task of labeling the data.
Although these crowdworkers individually may not be as accurate as an expert, constructing the true label from their aggregated opinions can approximate the accuracy of an expert.
However, the subjectivity of annotators and their different qualifications introduce noise to the labeling process.
To model this noise, most studies either focus on modeling the reliability of annotators and their correlation and reflecting it in the label aggregation phase or combining classifier training with learning the annotators' trust parameters.
Yet, learning through crowdsourcing-based models can still fail in the presence of differing annotations and unreliable users \cite{9288508}.

A promising direction for dealing with noisy labels for training complex classifiers is Co-teaching \cite{han2018co}.
Under the Co-teaching paradigm, two peer neural networks are trained separately and specific samples are exchanged between the networks to reduce the error of the two models and yield a more accurate model.
As a result, Co-teaching methods have shown great promise for computer vision problems with noisy labels.
Co-teaching can naturally counteract crowdsourcing noise since it filters out noisy samples in the beginning and only adds them at later training stages when they will be valuable.
However, Co-teaching treats each sample with the same weight.
This can cause the classifier to incorrectly learn from samples that may have fewer annotations or diverging human labels.

To address this limitation, we propose to leverage the certainty of samples from the label aggregation phase to inform the selection process of Co-teaching, which has not been studied before.
Our model, \method, uses a perturbation scheme based on the uncertainty of the samples to improve the robustness of the Co-teaching framework.
Given the availability of samples’ uncertainty from the label aggregation step, our model uses this information to counter the inherent noise by perturbing the input space.
In addition, the framework prioritizes the more confident samples of the classifier during the learning process. Thus, we tackle the problem of classification with features and crowdsourcing labels using three mechanisms:
\begin{description}

 \item[$\bullet$] Estimation of the features’ distributions to generate synthetic data which is then used to perturb each sample in an additive manner, proportional to its estimated label’s uncertainty.
 
 \item[$\bullet$] Enhancing Co-teaching by knowledge distillation, i.e. a student-teacher model of a simple and a complex network to accommodate smaller tabular data.

 \item[$\bullet$] Utilization of the perturbed samples as input to the above classifier to further differentiate uncertain and certain training points based on their loss in each epoch 
\end{description}

Next, we formally define the problem and summarize and delineate where and how \method~ties into the relevant literature in crowdsourcing, data augmentation, and learning with noisy labels.

\subsection{Problem Definition: Classification with Crowdsourcing Annotations}

In practice, there are numerous applications in which the ground truth of a classification task is not available, or disputed. For instance in medicine, multiple pathologists do not always necessarily agree on the malignancy status of a tumor in an image \cite{MobadersanyE2970}, or multiple nurses
do not all agree on the presence of hospital-acquired bedsores for a patient given their charts \cite{PUIagr}. Similarly, obtaining ground truth from experts to train reliable classifiers can be expensive, as in the case of content filtering and regulation of posts on social media, which are distributed among multiple non-expert annotators to obtain some good quality labels \cite{fbcontent}.
Formally, we define learning with crowdsourcing labels as follows:
\begin{definition}
 (Classification with Crowdsourcing Annotations) Consider a set of $R$ annotators labeling $N$ samples with $K$ possible classes. Given an answer matrix $\textbf{A} \in \mathbb{R}^{N \times R}$ where each element $a_{nr}$ indicates the label for sample $n$ provided by annotator $r$, and the training feature matrix $\textbf{X}_{tr} \in \mathbb{R}^{N \times M}$, the goal is to train a classifier that accurately predicts the true labels for the test data using only its feature matrix $\textbf{X}_{ts}$. 
\end{definition}

We use ${K}$ to denote number of classes. Simulated data from the synthesizer used for perturbation is shown by $\textbf{S}$ and the perturbed samples are denoted with $\widetilde{\textbf{X}_{tr}}$. The set of continuous and discrete features are shown by $F_{c}$ and $F_{d}$ respectively. Table \ref{tbl:notation} summarizes the notations used throughout this paper.
\begin{table}[!ht]
    \centering

    \caption{Summary of Notations.}
    \label{tbl:notation}

 \begin{tabular}{|l|l |}
 \hline
 \textbf{Symbol} & \textbf{Description}\\ \hline
  \multicolumn{1}{|c|}{N}&Number of Samples\\ \hline
     \multicolumn{1}{|c|}{R}&Number of Annotators\\ \hline
          \multicolumn{1}{|c|}{{K}}&Number of Classes\\ \hline
          \multicolumn{1}{|c|}{$\alpha$}&Perturbation Fraction\\ \hline
         \multicolumn{1}{|c|}{$\textbf{X}_{tr}$}& Training feature matrix\\ \hline
  \multicolumn{1}{|c|}{$\textbf{A}$}& Answer matrix of all annotators\\ \hline
  \multicolumn{1}{|c|}{$\textbf{S}$}& Synthetic feature matrix\\ \hline
      \multicolumn{1}{|c|}{$\widetilde{\textbf{X}_{tr}}$}& Perturbed training samples feature matrix\\ \hline
 \multicolumn{1}{|c|}{$F_{c}$}& Set  of continuous features\\ \hline
  \multicolumn{1}{|c|}{$F_{d}$}& Set  of all discrete features\\ \hline
 \multicolumn{1}{|c|}{$\mathbf{P}$}& Class probability matrix \\ \hline
  \multicolumn{1}{|c|}{${c_{i}}$}& Certainty of \textit{i}-th\\ 
  \hline
 \end{tabular}

\end{table}

\subsection{Related works}
\label{sec:related}
Classification with noisy answers or multiple crowdsourced labels overlaps with three other areas: learning with crowdsourcing labels, data augmentation and synthetic data generation for robust learning, and selective gradient propagation.
Here we summarize the three main high-level approaches for learning with multiple annotations.

\partitle{\textit{Sequential}} This approach first uses a truth inference method to estimate the ground truth for training samples. The estimated label is then used to train a classifier.

A recent survey extensively comparing these models has shown the overall efficiency and utility of the D\&S method \cite{surveycrowd}. 
Our proposed model falls into this category, however, we introduce ideas from the two other overlapping areas to further improve the predictive performance of this basic classifier.

\partitle{\textit{Simultaneous}} The second perspective jointly tackles the problem of learning classifier parameters and the estimated ground truth of the samples. Albarqouni et al. uses the Expectation-Maximization (EM) algorithm and Maximum a posteriori estimation to iteratively compute these two sets of parameters until convergence \cite{albarqouni2016aggnet}. Yet, this method is computationally challenging especially for more complex classifiers.

\partitle{\textit{Individual annotator’s label modeling}} The last set of research works entail learning a model for each individual labeler.
Dr. Net was proposed to learn a classifier to reproduce the labels of each annotator and is composed of two phases, individual annotator modeling and learning labelers’ averaging weights for the final prediction \cite{guan2017said}. 
To overcome the computational challenge of simultaneous learning and Dr. Net, multiple crowd-layer variants were introduced to remove the computational burden of the EM loop \cite{rodrigues2017deep}, by first estimating the ground truth of samples and then attempting to replicate the individual annotator’s labels using a very simple neural network.
Unfortunately, such models require significant samples to properly learn a robust classifier.

\subsubsection{Data Augmentation and Synthetic Data Generation for Robust Learning.}

To overcome the obstacle of noisy labels or features, perturbation schemes and data augmentations have been investigated. 
In computer vision, data augmentation is done by applying operations like cropping and rotation to combat potential mislabelled training data \cite{berthelot2019mixmatch,zhang2020distilling,soans2020sa}. 
Another line of work achieves robustness against noisy data by generating data synthesizers that achieves the same predictive performance as using the real data. Xu et al. have extended data augmentations to tabular data with heterogeneous feature types using Generative Adversarial Networks and Variational Autoencoders\cite{xu2019modeling}. 
However, such synthesizers are modeled independent of the labels or the conflicting annotations.

\subsubsection{Selective Gradient Propagation} 
To counter noisy labels and memorization effects in neural networks, the Co-teaching algorithm adaptively changes both the number of and the set of participating samples used in stochastic gradient descent epochs for two differently-initialized classifiers \cite{han2018co}. 
For each epoch, Co-teaching chooses a different number of samples with the lowest loss (as a proxy for clean data) and updates each classifier using the clean samples of the other network. This is in contrast to using all the samples or the clean samples of the classifier itself that may result in memorization and early overfitting which prohibits learning a generalizable classifier.
A parallel can be drawn to similarly deal with the inherent noisiness of aggregated crowdsourcing labels. Co-teaching mechanism of prioritizing a smaller set of confident samples in the initial stages of learning, and gradually incorporating more of the uncertain samples in later epochs can be leveraged for problem of classification with crowdsourcing labels.

%% file: method.tex
\section{Methodology}

Our idea is to enhance the Co-teaching framework to account for the uncertainty associated with the estimated truth label of the sample.
We introduce a perturbation-based scheme to the Co-teaching framework so the trained model will be more robust to sparsity and unreliability in the annotations. 
For each mini-batch update of Co-teaching, synthetic samples are generated and used to perturb each sample \emph{dependent} on the uncertainty of the estimated truth label. 
Thus a sample that has more certainty in the label will be perturbed more whereas a sample that has fewer annotations is likely to have less perturbation.
The perturbed sample is then used to train the classifier.
\begin{algorithm}[t]
\textbf{Input}: Training Features ${\textbf{X}_{tr}}$, Answer matrix $\textbf{A}$, Perturbation Fraction $\alpha$ \\ 
\textbf{Output}: $Model$\\
\SetAlgoLined
{Train synthesizer to create synthetic data: ${Data\_sampler}\xleftarrow[]{}Synthesizer(\textbf{X}_{tr})$}\\
Generate $N$ samples from resulting sampler: $\textbf{S}\xleftarrow[]{}Data\_sampler(N)$\\
{Run truth inference model to get class probabilities: $\textbf{P}\xleftarrow{}D\&S\_Algorithm(\textbf{A})$}\\
/* Generate perturbed samples $\widetilde{\textbf{X}_{tr}}$ */\\
\For{$i=1, \cdots, N$}{
Set sample's certainty using Eq. \eqref{eq:uncertainty}\\ 
Sample $s_i$ from 10\% closest samples of synthetic samples $\textbf{S}$ to ${x}_{i}$ using KNN \\
/* Generate continuous features */\\
\For{$j \in F_c$}{
    Generate feature $\widetilde{{x}}_{ij}$ according to Eq. \eqref{eq:contfeat} \\
}
/*Generate discrete features*/\\
    Calculate $f^{i}_d$ using Eq. \eqref{eq:f_d} \\
    Sample discrete features to perturb: $F^{i}_{d_{p}}$ from $F_d$ such that $|F^{i}_{d_{p}}| =  f^{i}_d$\\
    \For{$j \in F^{i}_{d_{p}}$}{
    Generate single feature value $\widetilde{{x}}_{ij}$ according to Eq. \eqref{eq:discfeat} \\
}
}
Train Co-teaching on Perturbed Samples: $Model\xleftarrow[]{}Co\_teaching(\widetilde{\textbf{X}_{tr}})$\\
 \caption{CrowdTeacher.}
 \label{alg:method}
\end{algorithm}

\subsection{Generating Synthetic Samples}

To improve the robustness of the Co-teaching framework, \method~generates synthetic samples of the data which are then used to perturb the samples to train the classifier.
Any data synthesizer with reasonable data generation performance can be used.
For the purpose of our paper, we focus on three data synthesizers: Conditional GAN (CTGAN) \cite{xu2019modeling}, TVAE \cite{xu2019modeling} and Gaussian copula \cite{sdv}. 
CTGAN can handle mixed feature types (discrete and continuous) and has been shown to perform competitively with other GAN-based, VAE-based, and Bayesian network-based data synthesizer for vision benchmark datasets \cite{sdv}.
It is worthwhile to note that the data synthesizer is not tied to the learning task and can be used as a stand-alone tool.

To generate synthetic data within \method, the training feature matrix $\textbf{X}_{tr}$ is fed to the synthesizer.
For CTGAN synthesizer, the discrete features $F_{d}$ are specified explicitly since they are modeled differently compared to the continuous features $F_{c}$.
Once the synthesizer has estimated the data distribution, any number of samples can be drawn.

For \method, we generate the synthetic set $\textbf{S} \in \mathbb{R}^{N \times M}$ with $N$ synthetic samples once and assume each synthetic sample can serve as a unique perturbation source.
Although $\textbf{S}$ is drawn once and is the same size as our training data to minimize the computational footprint of our model, the synthetic set can be re-drawn at each mini-batch of the Co-teaching framework with a larger number of samples.
\subsection{Sample-specific Perturbations}

The generated synthetic samples, $\textbf{S}$, fail to account for the uncertainty associated with the estimated sample label as the synthetic samples are only dependent on original training data.
Thus, we introduce a mechanism to leverage the uncertainty that arises from the truth inference method to individually perturb each sample.
For the purpose of illustration and experimentation, we focus on the D\&S algorithm \cite{{dawid1979maximum}}, but note that \method~can be used with any robust truth inference method that quantifies the label uncertainty for each sample.
The D\&S algorithm takes as an input the matrix of annotations ($\textbf{A}$) and models annotators by a confusion matrix to capture their chance of mistaking one class for another or correctly reporting them in addition to the class priors. D\&S outputs a matrix $\mathbf{P} \in \mathbb{R}^{N \times K}$, where the $P_{ik}$ element denotes the probability that sample $i$ is of class $k$. The certainty of each sample, $c_i$, is then defined as the maximum probability across all the classes:

\begin{equation}
\centering
\label{eq:uncertainty}
 c_{i} = \max_{k \in {K}}{(P_{ik})} \quad  \forall i \in N  
\end{equation}

\partitle{\textit{Choosing an appropriate simulated sample for perturbation}}
Given the data synthesizer can generate synthetic samples that are quite different from the original data point and can lead to more uncertainty with respect to the truth label, we use k-nearest neighbors (KNN) to identify reasonable close samples from $S$.
For each sample, KNN is run to find the top 10\% closely simulated samples. A simulated data point, $s_i$, is then randomly chosen from this top 10\% and used to perturb the original point.

\partitle{\textit{Perturbation}}
Each sample $x_i$ is perturbed using the simulated data point $s_i$ according to the uncertainty, $c_i$ and a user-specified perturbation fraction $\alpha \in [0, 1]$ to obtain the perturbed sample $\widetilde{{x}}_{i}$.

Let $s_{ij}$ represent the $j^{th}$ feature of sample $s_i$. 
If the $j^{th}$ feature is continuous, the value for the synthetic, perturbed sample $\widetilde{{x}}_{ij}$ is a convex combination of the original and simulated sample:
\begin{equation}
\centering
\label{eq:contfeat}
\widetilde{{x}}_{ij} = ({1-{\alpha} c_{i}}) {{x}_{ij}} + ({{\alpha} c_{i}}) {{s}_{ij}},  \quad \forall i \in N ,\quad \forall j \in {{F}_{c}}
\end{equation}

For the discrete features, we use $c_i$ and $\alpha$ to calculate the number of discrete features to swap.
Let $|F_d|$ denote the number of discrete features in the dataset, then the number of discrete features to swap for each sample $x_i$, $f^{i}_d$ is calculated as:
\begin{equation}
\centering
\label{eq:f_d}
f^{i}_d = round({\alpha}c_{i}|{F}_{d}|)
\end{equation}
Then $f^{i}_d$ features are randomly selected for perturbation from the original discrete feature set and denoted as $F^{i}_{d_{p}}$.
For each feature, $j$ in this perturbation set, the feature values are replaced with the synthetic sample value $s_{ij}$.

\begin{equation}
\centering
\label{eq:discfeat}
\widetilde{{x}}_{ij} = {{s}_{ij}},  \quad \forall i \in N ,\quad \forall j \in F^{i}_{d_{p}}
\end{equation}

\subsection{Knowledge distillation-based Co-teaching for Smaller Tabular Data}
To combat the large performance variations associated with running the Co-teaching algorithm on smaller-sized tabular data, we incorporated the student-teacher idea from knowledge distillation \cite{hinton2015distilling}.
Thus instead of two peer networks with the same architecture, we used one simple and one complex network such that the number of hidden units of the simpler network is half of the other one. 
Empirical results showed these modifications helped with both the convergence of the two networks in achieving more similar evaluation metrics and overall better performance across different synthetic datasets. 



%% file: experiment.tex
\section{Experiments}

\subsection{Baseline Methods}

The best performing methods from crowdsourcing studies (see Sec. \ref{sec:related}) 
are chosen as comparison models.
The original Co-teaching algorithm and Co-teaching using only uniformly perturbed input are also used to illustrate the advantage of certainty-aware perturbation.
All methods employ the same base classifier, a neural network with one hidden layer of $\frac{|{F}_{c}| + |{F}_{d}|}{4}$ units. 
Sequential methods share the same truth inference method (D\&S) and are marked with *. 
\begin{itemize}
    \item Naive baseline* (Base\_clf) \cite{dawid1979maximum}: Base classifier trained with D\&S labels.
    \item Simultaneous Expectation Maximization (S-EM) \cite{albarqouni2016aggnet}: An algorithm that jointly learns the classifier and annotators’ parameters using EM algorithm.
    \item Dr. Net \cite{guan2017said}: An individual annotation based model that separately learns each annotator’s labels and their weights.
    \item Crowdlayer (CL\_MW and CL\_VW) \cite{rodrigues2017deep}: An algorithm that estimates ground truth first and replicates each annotator’s labels via a simple final layer. This final layer is removed at test time. The number of parameters for the last layer determines the Crowdlayer variant. We evaluated the vector of weights (VW) and matrix of weights (MW) variants.
    \item Vanilla Co-teaching* (V\_Coteach) \cite{han2018co}: The original Co-teaching algorithm trained with D\&S labels.
    \item Co-teaching with uniform perturbation* (P\_Coteach): The Co-teaching algorithm trained on D\&S labels and synthetic samples.
    \item \method*: Our proposed method with the Co-teaching algorithm trained on D\&S labels and sample-specific certainty-informed perturbed samples.
\end{itemize}

We conducted our experiments using these baseline models. Since S-EM and Dr. Net constantly performed poorly compared to the other baselines, we omitted them from the plots for better readability. The Python implementation for all our experiments is publicly available on GitHub.
\subsection{Annotation Simulation}

For our experiments, we set the number of annotators to be 5 ($R=5$). 
To simulate the annotators' behavior, we consider two parameters: (1) mean reliability, or the average likelihood of the 
annotators to label a positive sample correctly and (2) variability in annotators’ expertise or the difference in their qualities.
We set the distribution of samples having 1 to 5 labels as [$\tau$ , 0.55*($1-\tau$), 0.27($1-\tau$), 0.13($1-\tau$), 0.05($1-\tau$)] and vary the parameter $\tau$ for our experiments. Note that $\tau$ determines the average number of labels per sample.
Conventionally, the Beta distribution is used to generate each annotator’s reliability. After determining each annotator’s reliability, its labels are created by randomly choosing (100-reliability) percent of positive cases and switching their labels into negative 0. Flipping negative samples to positive occurs at 0.01 times this rate. Samples not assigned to specific annotators are marked with $-1$ in the answer matrix ($\textbf{A}$). The exact parameters used for simulating annotations in each experiment are summarized in the GitHub repository.

\subsection{Datasets}

\subsubsection{Synthetic Datasets}

To test the performance of our framework on a non-specific dataset for which the ground truth is known, we generated synthetic data to mimic real-world features and a range of annotator reliabilities.

\emph{Statistical distribution families:}
Families of continuous and discrete distributions were used to generate the synthetic data. In particular, we used Normal, Beta, Wald, Laplace, Binomial, Multinomial, Geometric and Poisson distributions. The corresponding distribution parameters for a feature within each family are randomly chosen from a specified range. 5 features were chosen from each family for a total of 40 features.

\emph{Output:} The ground truth labels are determined based on a polynomial combination of feature values. Each feature's coefficient value is chosen randomly. 
To assign labels and model class balance (\% of positive samples), outputs falling in percentiles below the level of balancedness are assigned to the positive class. 

\emph{Noise level}: Two versions of labels are generated. Labels for a specified percentage of samples are flipped to obtain the noisy truth used for annotation generation. However the true labels before flipping are used for evaluation purposes. This resembles the availability of noisy labels in practice.

\subsubsection{PUI Dataset}

Determining whether a patient has developed a pressure ulcer injury (bedsore) is a complex clinical decision that requires considerable nursing expertise. Early detection of PUI is extremely useful since it is preventable with proper care. However, even highly trained nurses do not agree on the existence or severity of PUI cases. Training a classifier that utilizes a limited set of annotated health records from multiple nurses can revolutionize nursing care through use in similar clinical settings.
We use the MIMIC-III dataset \cite{origMIMIC}, a publicly available dataset which holds information of patients admitted to intensive care units (ICU) of a populated tertiary care hospital from 2001 to 2012. 
We identified hospital stays of individuals over 20 years old with length of stays between 2 days and 120 days.
A hospital stay was considered positive if there was a presence of the ICD-9 diagnosis code associated with pressure ulcer and there was a mention of PUI in the notes.
A hospital stay was negative if there was no indication of PUI in both the ICD-9 codes or the notes.

A total of 10518 samples were identified, 31\% of which are positive. 

%% file: result.tex
\section{Results}

Since the datasets are imbalanced, we evaluate all the models based on the area under the precision recall curve (AUPRC). 
AUPRC offers a holistic picture of \method's~ predictive performance, independent of the classification threshold choice. We split each dataset into 80\% training \& 20\% test. The AUPRCs in plots are averaged across multiple seeds. We also confirmed \method~performance on AUROC metric, but omit the results due to limited space.

\subsection{Synthetic Dataset}

\subsubsection{Sensitivity to choice of synthesizer}
To analyze the effect of using different synthesizers on \method~performance, we compared the average gain obtained by using \method~ with CTGAN, TVAE, and Gaussian copula synthesizers compared to using the next two top-performing baseline methods of P\_Coteach and V\_Coteach, respectively shown by circle and cross markers in Figure \ref{fig:syntcomp}. Firstly, we can see that Gaussian copula has the greatest gain among the three synthesizers. However, employing the two other synthesizers for \method~would still be beneficial in terms of predictive performance in many of the sparsity settings. Given the promising performance of Gaussian copula synthesizer, we use Gaussian copula for all the remaining experiments.

\subsubsection{Sensitivity to perturbation fraction ($\alpha$)}
\begin{figure}[ht!]
\begin{subfigure}{.5\textwidth}
\centering
    \caption{Perturbation fraction.}
    \label{fig:pfexp}
 \centering\includegraphics[scale=0.42]{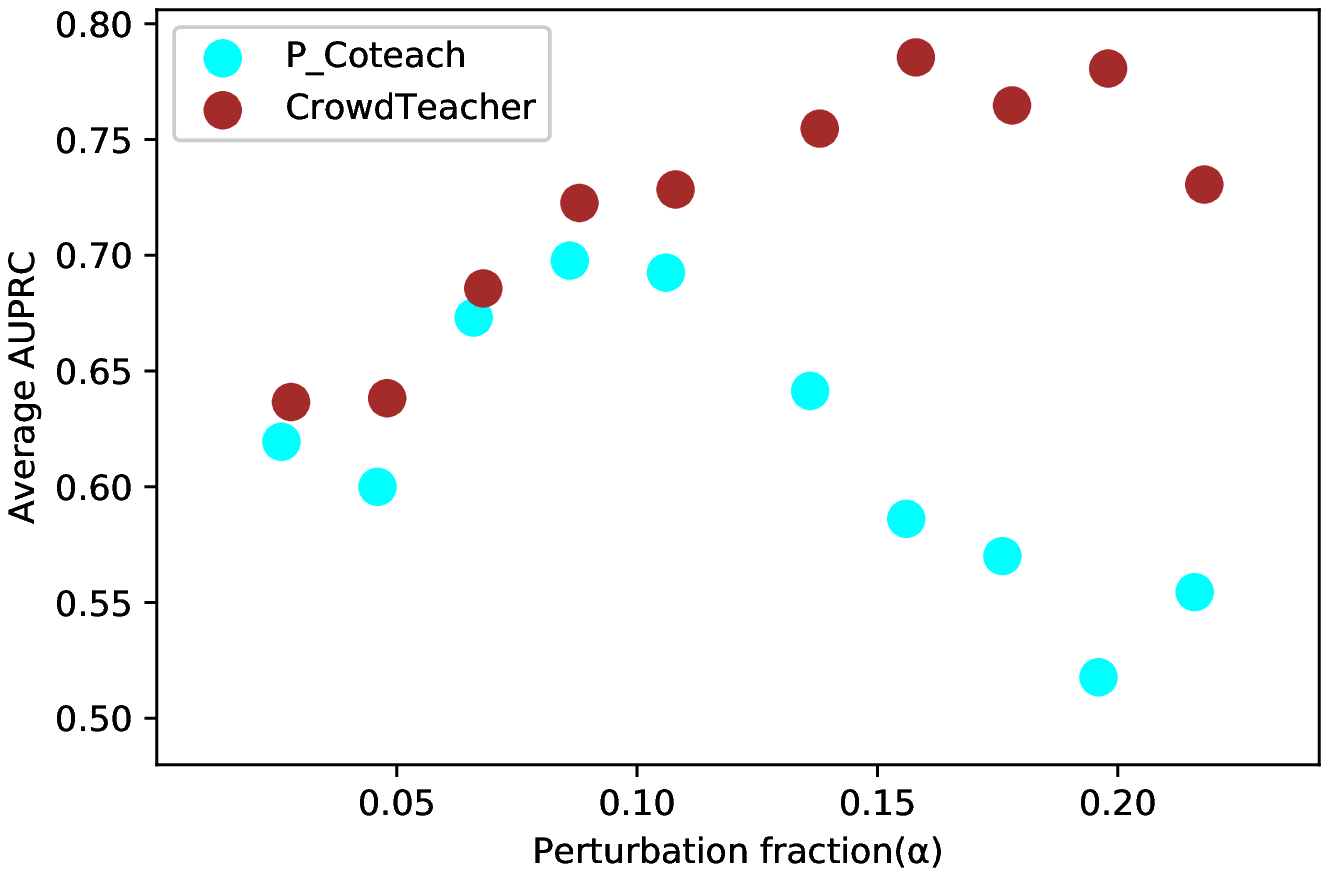}
\end{subfigure}%
\begin{subfigure}{.5\textwidth}
    \centering
    \caption{Different synthesizers.}
 \label{fig:syntcomp}
    {
 \centering\includegraphics[scale=0.42]{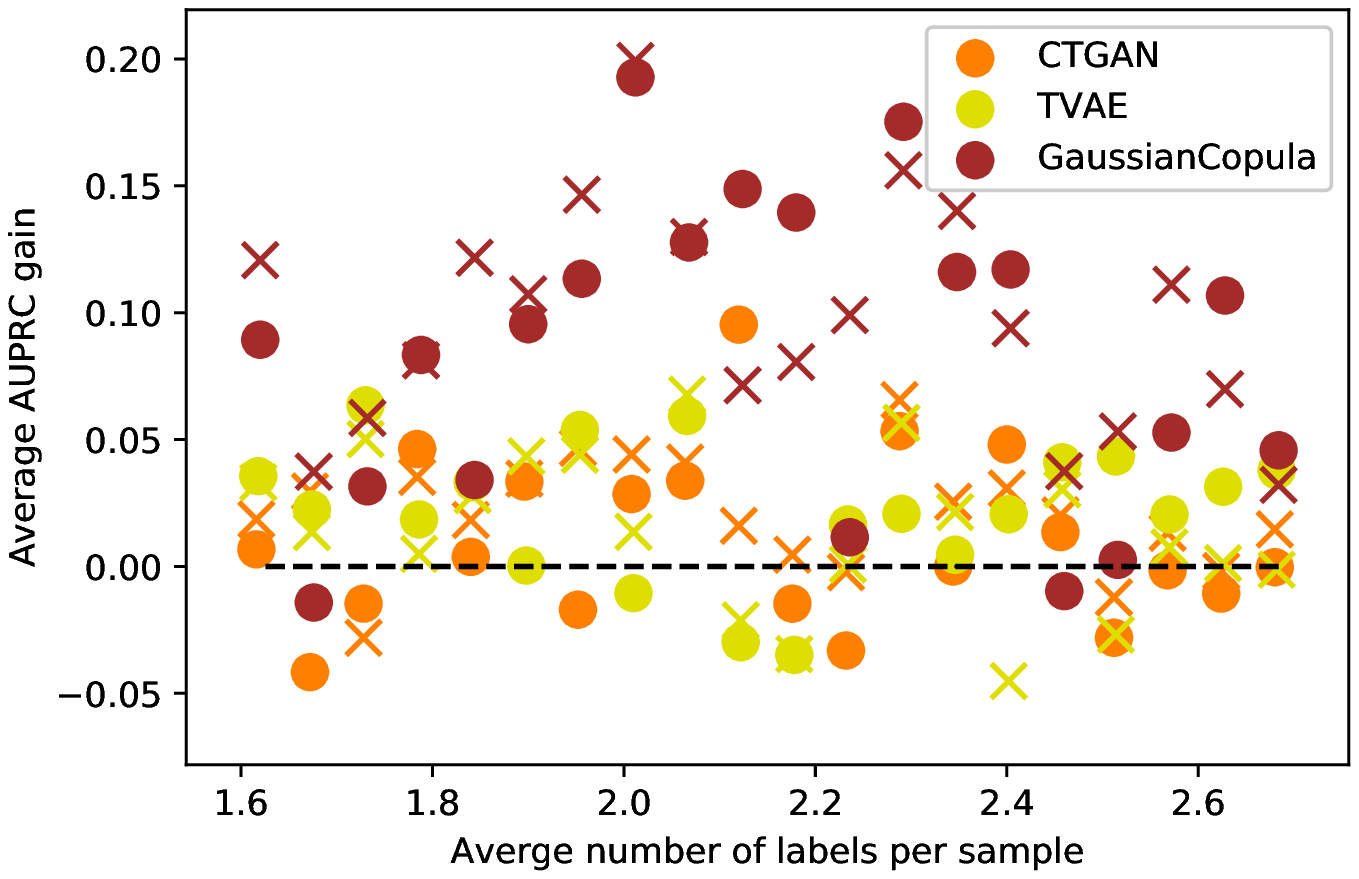} 
 }	
\end{subfigure}
\caption{CrowdTeacher Sensitivity to perturbation fraction and synthesizer choice (in Figure~\ref{fig:syntcomp} circles/crosses show gain w.r.t. P\_Coteach/V\_Coteach accordingly)}
\label{fig:sensitivity}
\end{figure}

To understand the impact of the perturbation fraction, $\alpha$, we varied it between [0.01, 0.2] and evaluated the performance of \method~and P\_Coteach (the two perturbation-based methods).
Figure \ref{fig:pfexp} shows the average AUPRC of P\_Coteach and \method~as $\alpha$ increases with the average number of labels set to 2.34. It is observed that \method~constantly outperforms P\_Coteach regardless of the chosen perturbation fraction indicating its robustness. From the results, there is an optimal range of $\alpha$ to achieve the greatest benefit from \method~and that either a very low ($\alpha \leq 0.05$) or very high ($\alpha \geq 0.2$) perturbation fraction decreases the usefulness of \method~but does not diminish it. Given these results, the remainder of our experiments uses $\alpha = 0.11$.

\subsubsection{Predictive Performance}
\begin{figure}[ht!]
\begin{subfigure}{.5\textwidth}
\centering
    \caption{Synthetic dataset.}
    \label{fig:synall}
 \centering\includegraphics[scale=0.42]{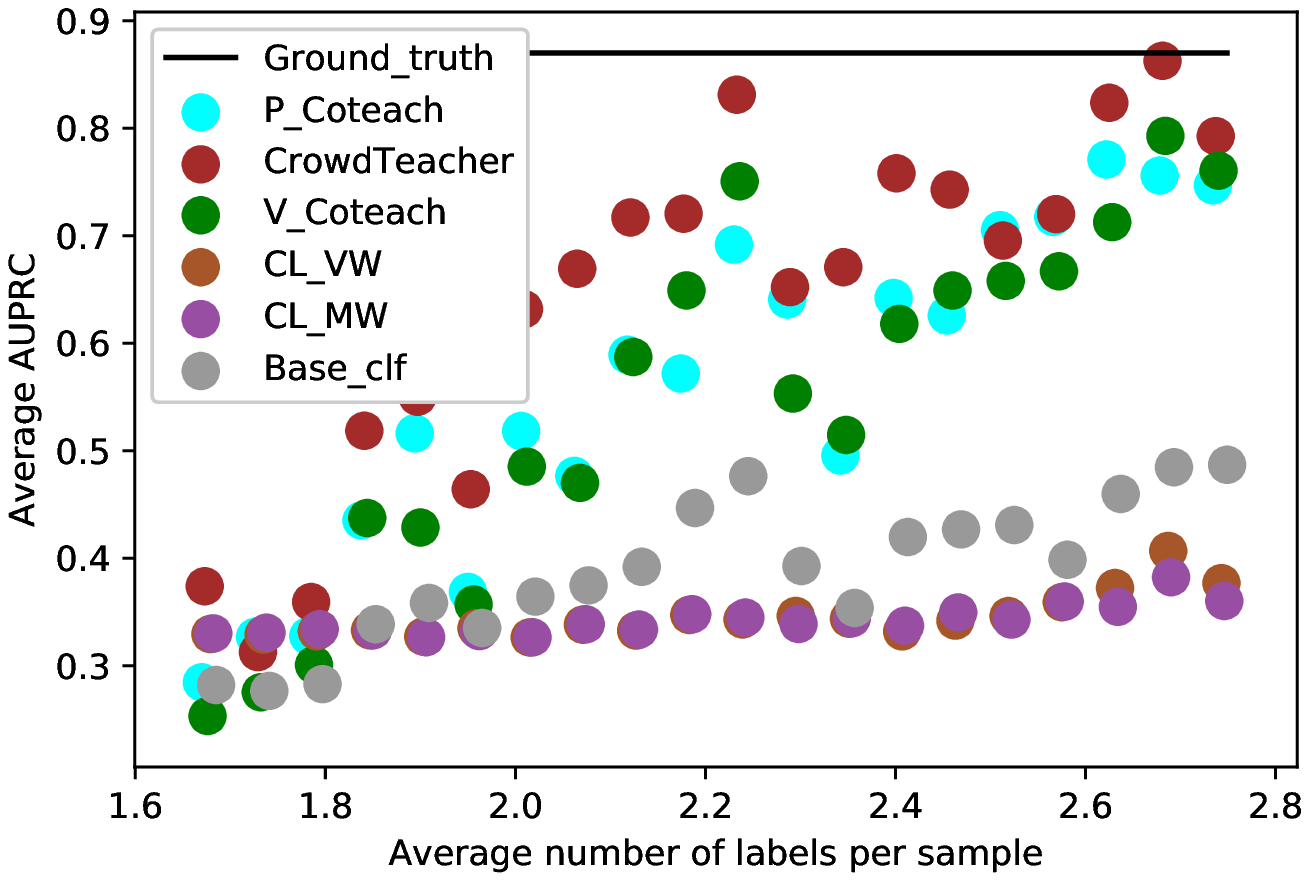}
\end{subfigure}%
\begin{subfigure}{.5\textwidth}
    \centering

    \caption{PUI dataset.}
 \label{fig:puiall}
    {
 \centering\includegraphics[scale=0.42]{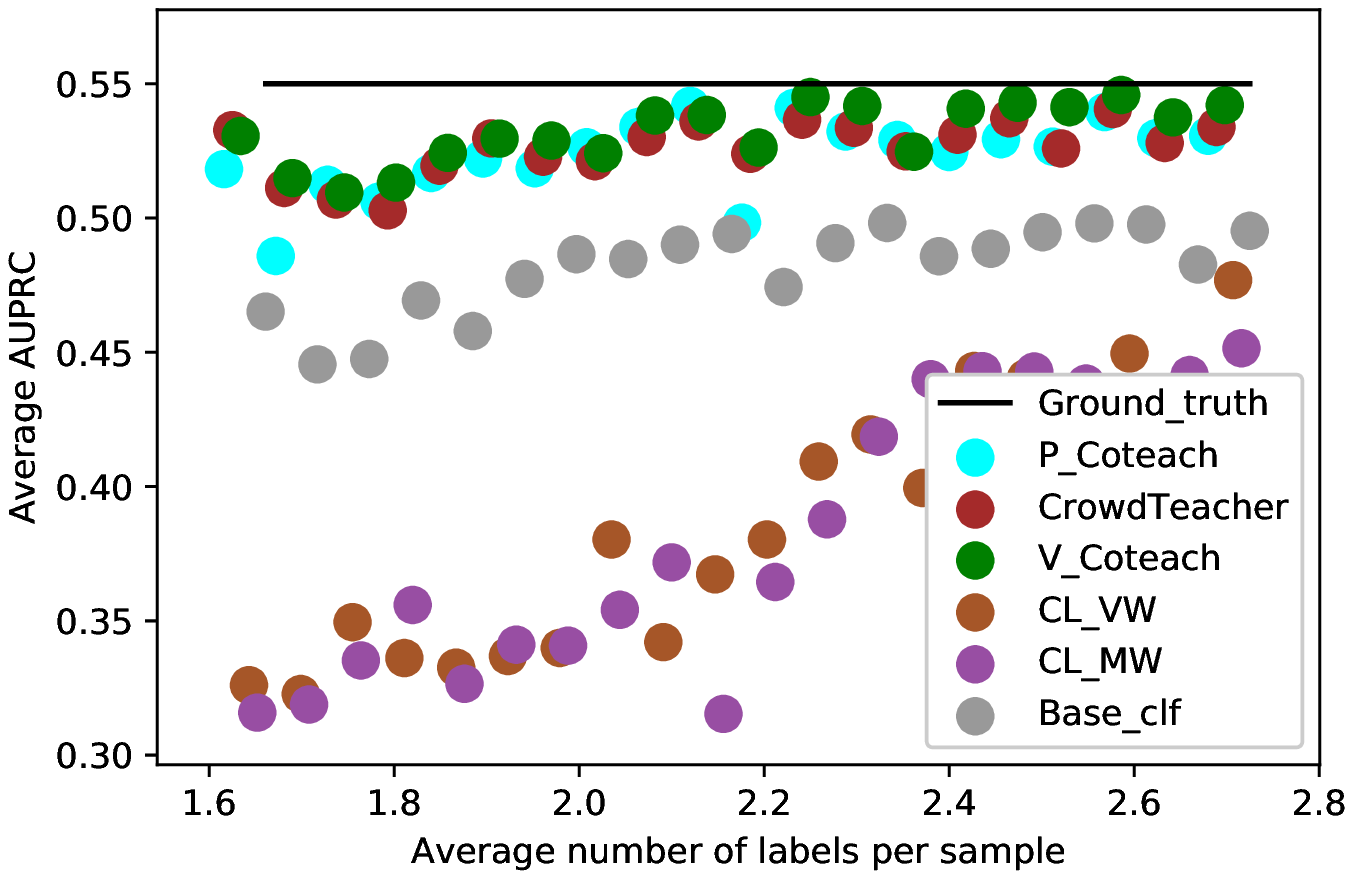} 
 }	
\end{subfigure}
\caption{CrowdTeacher Performance on Synthetic and PUI data as average number of labels per sample increases, averaged on 10 and 4 initializations respectively.}
\label{fig:exp}
\end{figure}

Figure \ref{fig:synall} shows the performance of baseline crowdsourcing and Co-teaching variants against CrowdTeacher across various sparsity settings on the synthetic dataset. Confirming intuition, all methods experience an increase in AUPRC since the average number of labels per sample increases, which exposes methods to less noisy annotation. All Co-teaching based methods (\method, V\_Coteach, and P\_Coteach) constantly outperform both crowdlayer variants and also Dr.Net and S-EM. The last two always performed the worst and therefore were excluded from these plots. Even though the base classifier performance improves with more labels, its performance gap with Co-teaching based methods remains large in all sparsity settings. Across a wide range of label sparsities, using \method~results in a significant boost in AUPRC, compared to the other two Co-teaching based methods, even with as low as only 1.68 labels per sample. Also, we can observe that V\_Coteach performs worse than P\_Coteach in very sparse settings (average number of labels $<$ 2.1), but as the number of labels increases it catches up with P\_Coteach and even surpasses it at higher densities. Another interesting observation is that beyond an average of 2 labels per sample, all three methods reach a plateau and only improve negligibly in response to an increased number of labels.

\subsection{PUI Dataset}

To challenge \method's performance under more chaotic distributions of real data, we tested it on the bedsore detection task with 10k samples. Figure \ref{fig:puiall} shows how the performance of the chosen methods changes as the average number of labels per sample goes up. We observed similar patterns to synthetic dataset here too in terms of Co-teaching variants' overall predictive advantage over other methods, however, the gap between Co-teaching variants and other methods is less substantial. The range of AUPRC of all models on this dataset proves that this is a much harder learning problem, yet \method~is able to beat P\_Coteach and V\_Coteach at multiple points, especially at lower sparsities, which are actually more practical for obtaining labels for hospital-acquired bedsores, while at other sparsity points it has comparable performance to these methods.

%% file: conclusion.tex
\section{Conclusion}

We proposed \method, a novel Co-teaching based approach that leverages certainty of samples from truth inference algorithms to apply sample-specific perturbations on training points, and combines it with Co-teaching algorithm to further rectify noisy annotations and incorporate that knowledge in the training process. Our proposed approach bridges overarching themes and ideas from data augmentation, crowdsourcing, and learning with noisy labels and is agnostic to the truth inference method and the synthesizer used. To illustrate the predictive benefits of CrowdTeacher over similar methods, we conducted experiments on both synthetic and real dataset of different scales, and our results for both tasks (including a real-world medical classification task) confirmed \method's performance edge for learning with crowdsourced labels.
We also successfully employed Co-teaching mechanism primarily tested on images, for tabular data. For our future work, we plan to propose new perturbation schemes to introduce more variety for perturbations of a given sample during training, and extend our current framework to semi-supervised learning.